\documentclass[runningheads]{llncs}

\usepackage{eccv}

\usepackage{eccvabbrv}

\usepackage{graphicx}
\usepackage{booktabs}

\usepackage[accsupp]{axessibility}  

\usepackage{hyperref}

\usepackage{orcidlink}
\usepackage{amsmath}
\DeclareMathOperator*{\argmax}{arg\,max}

\usepackage[capitalize]{cleveref}
\crefname{section}{Sec.}{Secs.}
\Crefname{section}{Section}{Sections}
\Crefname{table}{Table}{Tables}
\crefname{table}{Tab.}{Tabs.}
\newcommand{\B}{\bfseries}

\usepackage{pifont}
\newcommand{\xmark}{\ding{55}}%

\newcommand{\CAnote}[1]{{\color{brown}{\bf CA: }#1}}

\newcommand{\KG}[1]{{\color{blue}#1}} 
\newcommand{\CA}[1]{{\color{teal}#1}} 
\newcommand{\CAn}[1]{{\color{brown}#1}} 
\newcommand{\CAcr}[1]{{\color{teal}#1}} 
\newcommand{\PY}[1]{{\color{violet}#1}}

\renewcommand{\CA}[1]{{\color{black}#1}} 

\renewcommand{\CAnote}[1]{{}}
\renewcommand{\CAcr}[1]{{\color{black}#1}} 
\renewcommand{\KG}[1]{{\color{black}#1}} 
\renewcommand{\CAn}[1]{{\color{black}#1}}  
\renewcommand{\PY}[1]{{\color{black}#1}}

\newcommand{\sharedsymbol}{\ensuremath{^{*}}}

\begin{document}

\title{Action2Sound: Ambient-Aware Generation of  Action Sounds from Egocentric Videos} 

\titlerunning{Action2Sound}

\author{Changan Chen\inst{1}\sharedsymbol \and
Puyuan Peng\inst{1}\sharedsymbol \and
Ami Baid\inst{1}\and
Zihui Xue\inst{1}\and \\
Wei-Ning Hsu\inst{2}\and
David Harwath\inst{1}\and
Kristen Grauman\inst{1}
}

\authorrunning{Chen and Peng et al.}

\institute{
University of Texas at Austin \and FAIR, Meta
}

\maketitle
\begingroup
\renewcommand\thefootnote{\sharedsymbol}
\footnotetext{indicates equal contribution.}
\endgroup
\begingroup


\vspace{-0.3in}
\begin{abstract}
Generating realistic audio for human actions 
is important for many applications, such as creating sound effects for films or virtual reality games.
Existing approaches implicitly assume total correspondence between the video and audio during training, yet many sounds happen off-screen and have weak to no correspondence with the visuals---resulting in uncontrolled ambient sounds or hallucinations at test time. 
We propose a novel \emph{ambient-aware} audio generation model, AV-LDM.  We devise a novel audio-conditioning mechanism to learn to disentangle foreground action sounds from the ambient background sounds in in-the-wild training videos. Given a novel silent video, our model uses retrieval-augmented generation to create audio that matches the visual content both semantically and temporally.
We train and evaluate our model on two in-the-wild egocentric video datasets, Ego4D and EPIC-KITCHENS, and we introduce Ego4D-Sounds---1.2M curated clips with action-audio correspondence.
Our model outperforms an array of existing methods,
allows controllable generation of the ambient sound, and even shows promise for generalizing to computer graphics game clips.
Overall, our approach is the first to focus video-to-audio generation faithfully on the observed visual content despite training from uncurated clips with natural background sounds.


\vspace{-0.05in}

\end{abstract}
\vspace{-0.35in}
\section{Introduction}
\vspace{-0.05in}
\label{sec:intro}


As we interact with objects around us in our daily lives, our physical actions often produce sound, e.g., clicking on a mouse, closing a door, or cutting vegetables. The distinct characteristics of these \emph{action sounds} depend upon the type of action being performed, the shapes and materials of the objects being acted upon, the amount of force being applied, and so forth.
Vision not only captures \emph{what} physical interaction happens but also informs us \emph{when} the interaction happens, suggesting the possiblity of  synthesizing semantically plausible and temporally synchronous action sounds from silent videos alone.  This capability would accelerate many real-world applications, such as text-to-video generation, generating sound effects for films (Foley), or sound effect generation for virtual reality (VR) and video games.

\begin{figure}[t]
    \centering
    \includegraphics[width=1\linewidth]{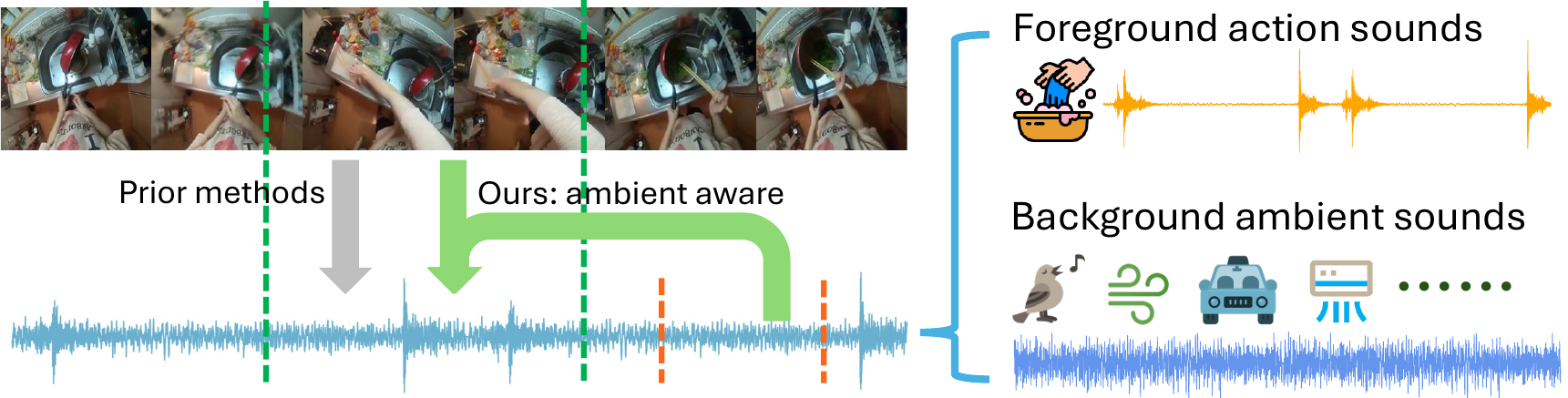}
    \vspace{-0.25in}
    \caption{Real-world audio 
    consists of both foreground action sounds (whose causes are visible) and background ambient sounds generated by sources offscreen. Whereas prior generation work 
    is agnostic to this division, 
    our method is ambient-aware and disentangles action sound from ambient sound. 
    Our key technical insight is how to train with in-the-wild videos exhibiting natural ambient sounds, while still learning to factor out their effects on generation.  The green arrows reference how we condition generation on sound from a related, but time-distinct, video clip to achieve this.}
    \vspace{-0.25in}
    \label{fig:concept}
\end{figure}

Some prior work studies impact sound synthesis from videos~\cite{owens2016visually,Su2023PhysicsDrivenDM} while others target more general video-to-audio generation~\cite{iashin21taming,luo23difffoley}.
These methods \emph{implicitly assume total correspondence between the video and audio} and aim to generate the whole target audio from the video.
However, this strategy falls short
for in-the-wild training videos, which are rife with off-screen ambient sounds, e.g., 
traffic noise, 
people talking, or A/C running. While some of these ambient sounds are weakly correlated with the visual scene, such as the wind blowing in an outdoor environment, many of them have no visual correspondence, such as off-screen speech or a stationary buzzing noise from the fridge.
\CAcr{Most} existing methods are not able to disentangle action sounds from ambient sounds and treat them as a whole,
leading to uncontrolled generation of ambient sounds at test time and sometimes even hallucination, e.g., random action or ambient sounds.  
This is particularly problematic for generating action sounds because they are often subtle and transient compared to ambient sounds.  For example, trained in the traditional way, a model given a scene that looks like a noisy restaurant risks generating ``restaurant-like'' ambient sounds, while ignoring the actual movements and activities of the foreground actions, such as a person stirring their coffee with a metal spoon. 

How can we disentangle the foreground action sounds from background ambient sounds for in-the-wild video data \emph{without} ground truth separated streams? 
Simply applying a noise removal algorithm on the target audio does not work well since in-the-wild blind source separation of general sounds from a single microphone is still an open challenge~\cite{separation_review}, \KG{and class-dependent models for predicting visually relevant sounds cannot generalize to in-the-wild video~\cite{chen2020regnet}.}

Our key observation is that while action sounds are highly localized in time, ambient sounds tend to persist across time. 
Given this observation, we propose a simple but effective solution to disentangle ambient and action sounds: during training, in addition to the input video clip, we also condition the generation model on an audio clip from the same long video as the input video clip but from different timestamps. See \cref{fig:concept}. By doing so, we lift the burden of generating energy-dominating ambient sounds and encourage the model to focus on learning action cues from the visual frames to generate action sounds.
At test time, we do not assume access to (even other clips of) the ground truth video/audio.  Instead, we propose to retrieve an audio segment from the training set with an audio-visual similarity scoring model, inspired by recent ideas in retrieval-augmented generation (RAG)~\cite{Khandelwal2019GeneralizationTM,Guu2020REALMRL,Lewis2020RetrievalAugmentedGF}. This benefits examples where the visual scene 
has a weak correlation with the ambient sound that is 
beneficial to capture, e.g., outdoor environments.

Existing action sound generation work relies on either clean, manually-collected data that has a limited number of action categories~\cite{owens2016visually,Su2023PhysicsDrivenDM,clarke2021diffimpact}, or videos crawled from YouTube based on predefined taxonomies~\cite{gemmeke17audioset,chen20vggsound,iashin21taming,chen2020regnet}. 
To expand the boundary of action sound generation to in-the-wild human actions, we
take advantage of recent large-scale egocentric video datasets~\cite{grauman22ego4d,damen18scaling}. Though our model is not tailored to egocentric video in any way, there are two main benefits of using these datasets: 1) egocentric videos provide a close view of human actions compared to exocentric videos, where hand-object interactions are observed from a distance and are often occluded, and 2) these datasets have timestamped narrations describing atomic actions. We design a pipeline to extract and process clips from Ego4D, curating \textbf{Ego4D-Sounds} with 1.2 million audio-visual 
action clips.\footnote{\url{https://ego4dsounds.github.io}}

Our idea of disentangling action and ambient sounds implicitly in training is model-agnostic. In this paper, we instantiate it by designing an
audio-visual latent diffusion model (AV-LDM) that conditions on both modality streams for audio generation. We evaluate our AV-LDM against recent work on a wide variety of metrics 
and show that our model outperforms the existing methods significantly on both Ego4D-Sounds and EPIC-KITCHENS. We conduct a human evaluation study that shows our model synthesizes plausible action sounds according to the video. \textbf{Please see/listen for yourself in our supplementary video!}  
We also show promising preliminary results on virtual reality game clips.
To the best of our knowledge, this is the first work that demonstrates the disentanglement of foreground action sounds from background sounds for action-to-sound generation on in-the-wild videos.
\vspace{-0.15in}
\section{Related Work}
\label{sec:related_work}
\vspace{-0.1in}
\subsection{Action Sound Generation}
\vspace{-0.05in}
A pioneering work \KG{for capturing human-generated action sounds} 
collects videos where people hit, scratch, or prod objects with a drumstick~\cite{owens2016visually}.  \KG{This is an early inspirational effort, though it} is by design limited in the type of actions. The robotics community also studies this problem by using robotic platforms to collect collision sounds and analyze or synthesize them from video~\cite{Gandhi2020swoosh,clarke2021diffimpact}. Other work simulates collision events~\cite{gan21threedworld}, which remains difficult for action sounds due to the complexity of the physical interactions. 
\CA{\CAcr{Most} existing methods demonstrate good synthesis results 
when the data are noise-free. However, they 
\KG{are not equipped to} learn from in-the-wild action videos, where the action sound is always coupled with ambient sound. 
\CAcr{Sharing our motivation to disregard irrelevant sounds, the REGNET framework~\cite{chen2020regnet} aims to predict visually relevant sounds by conditioning on ground truth audio with a bottleneck design. 
 However, it does not allow controllable generation and risks learning to copy the target action and ambient sound leading to weaker empirical performance. \CAcr{More importantly, REGNET~\cite{chen2020regnet} requires curated datasets to train class-dependent models, which prevents generalization to in-the-wild data, as we will see in results.}}
We propose an ambient-aware model to deal with this issue \KG{head-on} and also introduce the Ego4D-Sounds dataset to expand action sound synthesis to in-the-wild actions.
}

\vspace{-0.15in}
\subsection{Egocentric Video Understanding \KG{with Audio}}
\vspace{-0.05in}
Understanding human activities in videos has long been a core challenge of computer vision. Early research studies activity recognition from exocentric video~\cite{ucf,kinetics,caba2015activitynet}.
Recent work  \CA{explores the egocentric setting and introduces} 
 large egocentric datasets such as Ego4D~\cite{grauman22ego4d}, EPIC-KITCHENS~\cite{damen18scaling}, and Ego-Exo4D~\cite{grauman24egoexo4d}. 
Leveraging both the video and audio streams in egocentric videos, many 
interesting tasks are enhanced, such as action recognition~\cite{kazakos19epicfusion},
localization~\cite{ramazanova2022owl}, active speaker localization~\cite{jiang2022egocentric,sagnik-cvpr2024}, sounding object localization~\cite{ego_av_localization,chen22soundspaces2}, 
and state-aware representations~\cite{mittal2022learning}. 
Most related to our work is SoundingActions~\cite{SoundingActions} which learns visual representations of actions that make sounds, \CAn{and is \KG{valuable for indexing and recognition problem settings, but ill-equipped for} generation, as we show later.}
All existing audio-visual learning for egocentric video focuses on perception, i.e., understanding what happens in the video.  In contrast, we target the video-to-audio generation problem. \KG{Furthermore, relative to any of the above, our idea 
to implicitly learn to disentangle the action sound from ambient sounds is novel.}

\vspace{-0.15in}
\subsection{Diffusion Models and Conditional Audio Generation}
\vspace{-0.05in}
Diffusion models have attracted significant attention recently because of their high fidelity generation~\cite{Dhariwal2021DiffusionMB,Nichol2021GLIDETP,Saharia2022PhotorealisticTD}. Initially proposed for image generation~\cite{Song2019GenerativeMB,ho20denoising}, 
they have also been successfully applied to speech and audio generation~\cite{Kong2020DiffWaveAV,Popov2021GradTTSAD,Yang2022DiffsoundDD,Liu2023AudioLDMTG,Huang2023MakeAnAudioTG}. Benefitting from classifier-free guidance~\cite{ho22classifierfree} and large-scale representation learning, AudioLDM~\cite{Liu2023AudioLDMTG} and Make-An-Audio~\cite{Huang2023MakeAnAudioTG} perform diffusion-based \emph{text-to-audio} generation. 
More recently, Diff-Foley~\cite{luo23difffoley} adapts latent diffusion models for video-to-audio generation by first conducting audio-video contrastive learning and then video-conditioned audio generation. While promising, it does not \KG{address} 
the background ambient sound problem. 
Inspired by recent work on retrieval-augmented generation (RAG) for text~\cite{Khandelwal2019GeneralizationTM,Guu2020REALMRL,Lewis2020RetrievalAugmentedGF,Borgeaud2021ImprovingLM} and image generation~\cite{Chen2022ReImagenRT,Blattmann2022RetrievalAugmentedDM}, \KG{we show our audio-conditioning insight carries over to inference time via a retrieval component of the model.}
Conditional video-to-audio generation conditions on either a physics prior to guide diffusion-based impact sound generation~\cite{Su2023PhysicsDrivenDM} or, in  CondFoleyGen~\cite{du23conditional}, another \PY{video clip}
to 
modify characteristics of the action sound. Our method also considers
additional conditioning signals to control the output, \KG{but for a very different purpose; our model
is the first to 
address foreground/background sound disentanglement in generation.}
\vspace{-0.1in}
\section{Ambient-aware Action Sound Generation}
\label{sec:approach}
\vspace{-0.05in}

\begin{figure}[t]
    \centering
    \includegraphics[width=\linewidth]{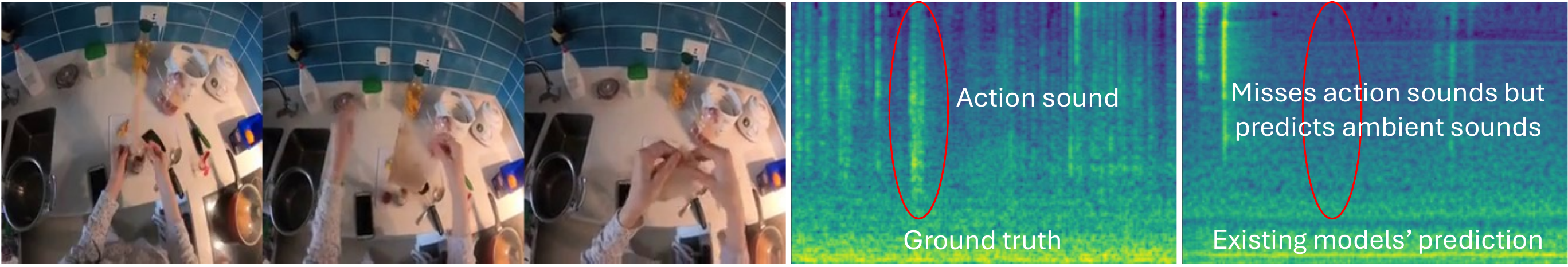}
    \vspace{-0.25in}
    \caption{Illustration of the harm of ambient sound in video-to-audio generation. 
    In this example, this person is closing a packet of ginger powder, which makes some rustling sound (red circled in the middle). There is also some buzzing sound semantically irrelevant to the visual scene in the background, which dominates the energy of the spectrogram. On the right-hand side, we show a prediction made by a vanilla model that misses the action sound but predicts the ambient sound.}
    \vspace{-0.15in}
    \label{fig:ambient_illustration}
\end{figure}

We first discuss our high-level idea of how to guide the generation model to disentangle action sounds from ambient sounds. We then devise AV-LDM, an extension of latent diffusion models (LDM) to accommodate both audio and video conditions. We also discuss our pretraining stage.

\vspace{-0.15in}
\subsection{Action-to-Sound Generation}
\vspace{-0.05in}
Given a video $V  \in \mathbb{R}^{(T*S_V)\times H\times W\times3}$, where $T$ is the duration of the video and $S_V$ is the video sample rate, and the accompanying audio waveform $A \in \mathbb{R}^{1\times (T*S_A)}$, where $S_A$ is the audio sample rate, 
our goal is to model the conditional distribution $p(A|V)$ for video-to-audio generation.
During training we observe natural video coupled with its audio, whereas at inference time we have only a silent video---e.g., could be an output from text-to-video, or a VR/video game clip, or simply a real-world video for which we want to generate new plausible sounds.

\vspace{-0.15in}
\subsection{Disentangling Action and Ambient Sounds}
\vspace{-0.05in}

Learning a video-to-audio generation model using in-the-wild egocentric videos is challenging because of entangled foreground action and background ambient sounds, as illustrated in \cref{fig:ambient_illustration}. More specifically, the reasons are two-fold: 1) while action sounds are usually of very short duration, ambient sounds can last the entire clip, and therefore dominate the loss, leading to low-quality action sound generation; 2) while some ambient sounds might be semantically related to the visual scene such as bird chirping in the woods, in many cases, ambient sounds are difficult to infer from the visual scene because they are the results of the use of certain microphones, recording conditions, people speaking, off-screen actions, etc. Forcing a generation model to learn those background sounds from video results in hallucinations during inference (see examples in \cref{fig:main_qual}).  

Therefore, it is important to proactively disentangle action sounds and ambient sounds during training. However, separating in-the-wild ambient sounds is still an open challenge: recent models rely on supervised training using artificially mixed sounds, for which the ground truth complex masks can be obtained~\cite{separation_review}. 
Simply applying off-the-shelf noise reduction methods to training data leads to poor performance, as we will show in \cref{sec:experiments}. 

While it is difficult to \emph{explicitly} separate the ambient and action sound in the target audio, 
our key observation is that ambient sounds are usually fairly stationary across time. 
Given this observation, we propose a simple but effective method to achieve the disentanglement.  During training, in addition to video clip $V$, we also provide the model an audio clip $A_n$ that comes from the same training video but a different timestamp as the input video clip (see \cref{fig:model}). Therefore, instead of modeling $p(A|V)$, we model $p(A|V,A_n)$. Given the hypothesis that $A_n$ is likely to share ambient sound characteristics with $A$, it can take away the burden of learning weakly correlated or even uncorrelated ambient sounds from visual input alone, and encourages the model to focus on learning action features from the visual input.
For the selection of $A_n$, we 
randomly sample one audio clip from the nearest $X$ clips in time. 
While there is no guarantee that the sampled audio shares exactly the same ambient sound with the target audio, their ambient sounds should largely overlap since they are close in time, which provides a consistent learning signal to help the model learn the disentanglement.
\CAcr{While is possible for the 
sampled audio to contain repetitions of the target action sound, 1) the chance of selecting a semantically relevant sound low (9\% based on (verb,noun) taxonomy) and 2) the precise temporal onset is almost never the same, thus making it impossible for the model to cheat in training.}
\begin{figure}[t]
    \centering
    \includegraphics[width=\linewidth]{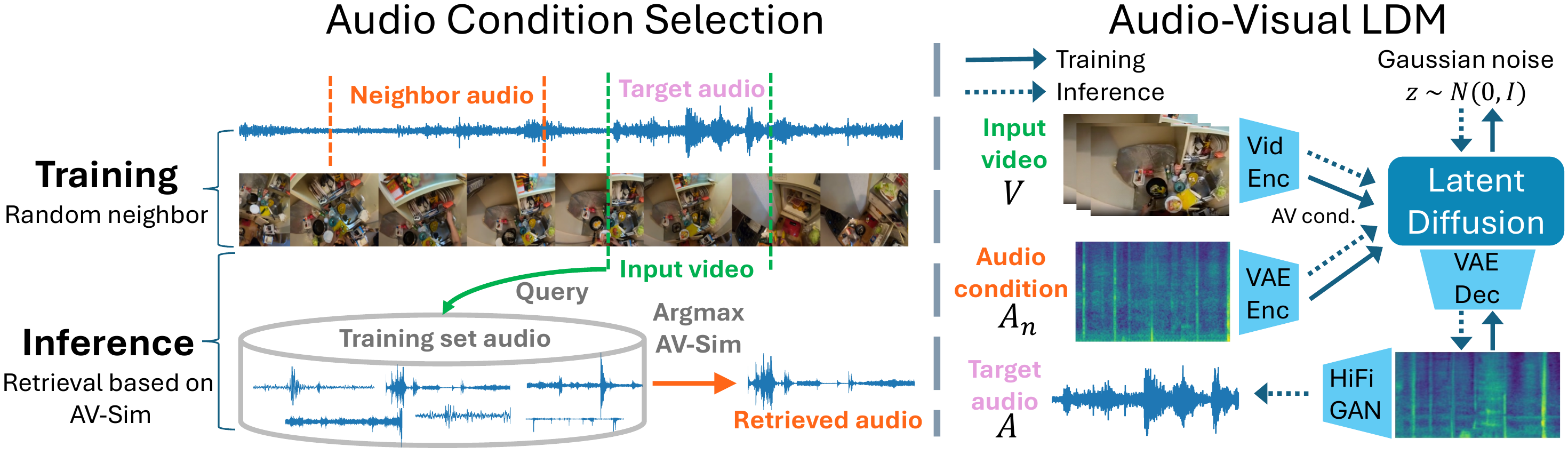}
    \vspace{-0.2in}
    \caption{Audio condition selection and the model architecture.
    \textbf{Left}: During training, we randomly sample a neighbor audio clip as the audio condition. For inference, we query the training set audio with the (silent) input video and retrieve an audio clip that has the highest audio-visual similarity with the input video using our trained AV-Sim model (\cref{sec:pretrain}).
    \textbf{Right}: We represent audio waveforms as spectrograms and use a latent diffusion model to generate the spectrogram conditioned on both the input video and the audio condition. At test time, we use a trained vocoder network to transform the spectrogram to a waveform.
    }\label{fig:model}
    \vspace{-0.2in}
\end{figure}

\vspace{-0.2in}
\subsection{Retrieval Augmented Generation and Controllable Generation}\label{sec:rag}
\vspace{-0.05in}
While during training we have access to the clips in the same long video as the input clip, we 
of course cannot access that  information at test time. How we select $A_n$ at test time depends on the purpose of the generation. We consider two use cases: \textit{action-ambient joint} generation and \textit{action-focused} generation. In the first scenario, we would like the model to generate both the action sound and the ambient sound that is plausible for the visual environment. This is useful, for example, for generating sound effects for videos. In the latter scenario, we would like the model to focus the generation on action sounds and \emph{minimize} ambient sounds, which is useful, for example, for generating sounds for games. \cref{fig:generation_scenarios} depicts the two scenarios.

For action-ambient joint generation, we want $A_n$ to be semantically relevant to the visual scene. Inspired by recent work in retrieval augmented regeneration, we propose to retrieve audio such that:
\setlength{\abovedisplayskip}{3pt}
\setlength{\belowdisplayskip}{3pt}
\begin{equation}
    A_n = \argmax _{A_i \in \mathcal{D}}{\text{AV-Sim}(A_i, V)}, 
\end{equation}
where $\mathcal{D}$ is the dataset of all training audio clips and $V$ is the (silent) input video. $\text{AV-Sim}(A, V)$ is a similarity scoring function that measures the similarity between $A$ and $V$, which we will cover in \cref{sec:pretrain}.

For action-focused generation, we want $A_n$ to have minimal ambient level. We find simply filling $A_n$ with all zeros results in poor performance, likely because it is too far out of the training distribution. Instead, we find conditioning the generation on a low-ambient sound will cue the model to focus on action sound generation and generate minimal ambient sound. See \cref{sec:ambient_control}.

\begin{figure}[t]
    \centering
    \includegraphics[width=0.6\linewidth]{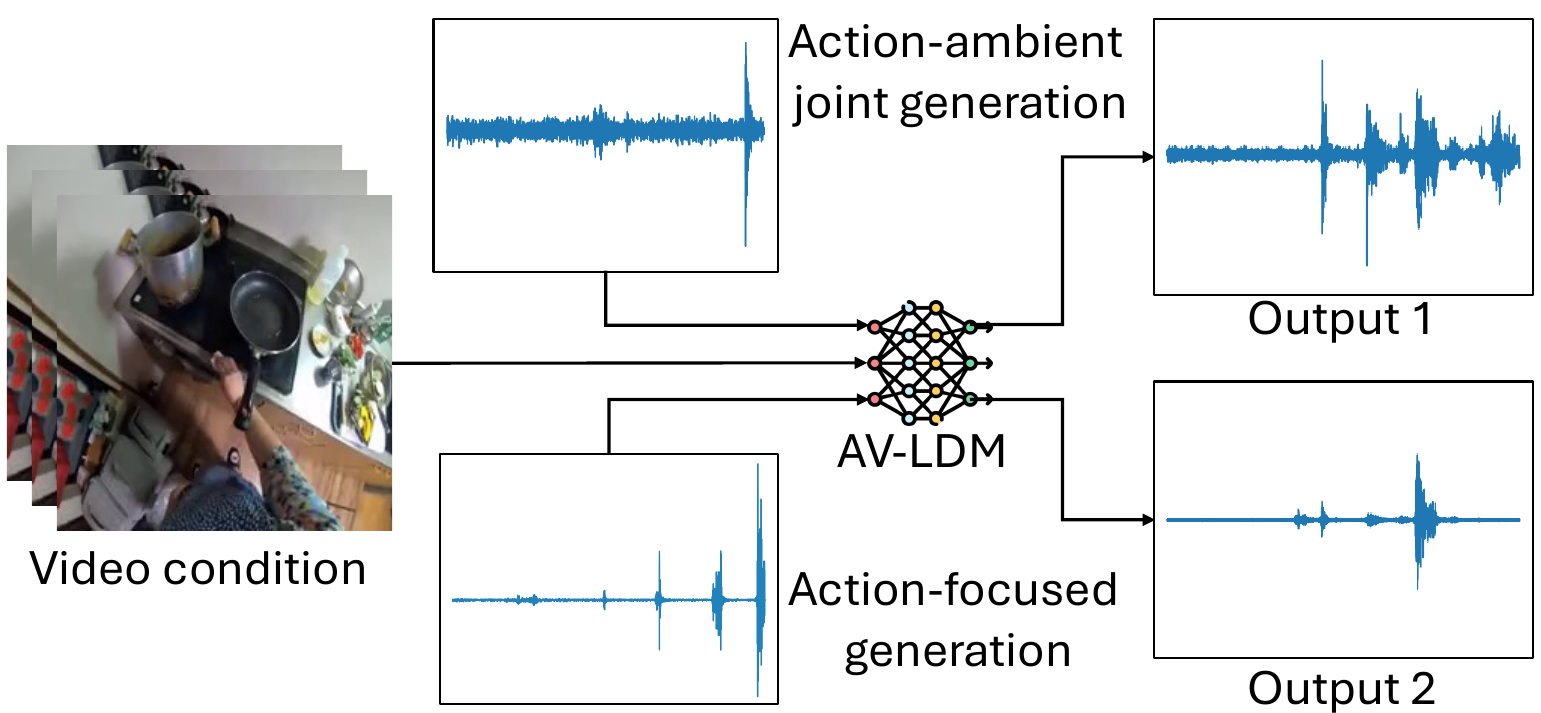}
    \vspace{-0.1in}
    \caption{Two inference settings: ``action-ambient joint generation'' and ``action-focused generation''. In the first setting, we condition on audio retrieved from the training set and aim to generate both plausible action and ambient sounds. In the second setting, we condition on an audio file with low ambient sound and the model focuses on generating plausible action sounds while minimizing the ambient sounds.
    }
    \label{fig:generation_scenarios}
    \vspace{-0.2in}
\end{figure}

\vspace{-0.15in}
\subsection{Audio-Visual Latent Diffusion Model}
\vspace{-0.05in}
While the above idea of disentanglement is universal and not specific to any model architecture, here we instantiate this idea on diffusion models due to their success in audio generation~\cite{Liu2023AudioLDMTG,luo23difffoley}. We extend the latent diffusion model to accommodate our audio-visual conditions, thus yielding an audio-visual latent diffusion model (AV-LDM).

\cref{fig:model} (right) shows the architecture of our model. During training, given audio waveform target $A$, we first compute the mel-spectrogram $x_0 \in \mathbb{R}^{T\times D_{\text{mel}}}$, where $D_{\text{mel}}$ is the number of mel bins. We then use a pretrained Variational Autoencoder (VAE) to compress the mel-spectrogram $x_0$ to a latent representation $z_0 \in \mathbb{R}^{C'\times H' \times W'}$, where $z_0$ is the generation target of the LDM. We condition the generation on both the video feature $c_v \in \mathbb{R}^{T_v,D_c}$ and audio feature $c_a \in \mathbb{R}^{T_a,D_c}$.
We extract the video feature with a pretrained video encoder (see \cref{sec:pretrain}) from $V$. We extract the audio feature from the audio condition $A_n$ with the same VAE encoder and then transform the feature into 1-d vector with a multilayer perceptron (MLP). 

Following~\cite{luo23difffoley}, we use cross attention where the query is produced by $z_t$, which is the sample diffusion step $t$, and key and value are produced by \\
$\text{concat}([\text{Pos}_v + c_v; \text{Pos}_a + c_a])$, where Pos denotes learnable positional embeddings.
The model is trained with the denoising objective:
\begin{equation*}
\mathcal{L} = \mathbb{E}_{t \sim \text{uniform}(1, T), z_0, \epsilon_t} \|\epsilon_t - \epsilon_\theta(\mathbf{x}_t, t, c_v, c_a)\|^2,   
\end{equation*}
where $\epsilon_t$ is the standard Gaussian noise sampled for diffusion step $t$, and $\epsilon_\theta(\mathbf{x}_t, t, c_v, c_a)$ is the model estimation of it ($\theta$ represents model parameters).

The reverse process can be parameterized as:
\begin{align*}
    &p(z_T) = \mathcal{N}(0,I), \\ 
    &p_{\theta}(z_{t-1}|z_t) = \mathcal{N}(z_{t-1}; \frac{1}{\sqrt{\alpha_t}} \Big(z_t - \frac{1 - \alpha_t}{\sqrt{1 - \bar{\alpha}_t}} \epsilon_\theta(z_t, t, c_v, c_a) \Big), \sigma_t^2I), 
\end{align*}
where $\alpha_t$ and $\sigma_t$ are determined by noise schedule of the diffusion process. To generate audio during inference, we first sample standard Gaussian noise $z_T$, and then apply classifier free guidance~\cite{ho22classifierfree} to estimate $\Tilde{\epsilon}_{\theta}$ as
\begin{equation*}
    \Tilde{\epsilon}_t(z_t, t, c_v, c_a) = \omega\epsilon_\theta(z_t, t, c_v, c_a) + (1-\omega) \epsilon_\theta(z_t, t, \emptyset, \emptyset),
\end{equation*}
where $\emptyset$ denotes zero tensor. For the above estimation to be more precise, during training, we randomly replace $c_v$ with $\emptyset$ with probability $0.2$. As for $c_a$, we found dropping it even with even a small probability harms the performance,
and therefore we always condition the LDM with $c_a$.

During inference, we use DPM-Solver~\cite{lu2022dpm} on LDM to sample a latent representation,
which is then upsampled into a mel-spectrogram by the decoder of VAE. Lastly, we use a vocoder (HiFi-GAN~\cite{kong2020hifi}) model to generate waveform from the mel-spectrogram.

\vspace{-0.2in}
\subsection{Audio-Visual Representation Learning}\label{sec:pretrain}
\vspace{-0.05in}
Generating semantically and temporally synchronized action sounds from video requires the video encoder to capture these relevant features. In addition, we would like to train a video model and an audio model whose representations align in the embedding space to support retrieval-augmented generation discussed in \cref{sec:rag}. For this purpose, we train a video encoder and audio encoder contrastively to optimize the following objective:
\begin{equation*}
    \text{AV-Sim}(A,V) = -\frac{1}{|\mathcal{B}|} \sum_{t \in \mathcal{B}}\log \frac{\exp(e_A^t  e_V^t/\tau)}{\sum_{l \in \mathcal{B}}\exp(e_A^t e_V^l/\tau)},
\end{equation*}
where $\mathcal{B}$ is the current batch of data, $e_A^t$ and $e_V^t$ are normalized embeddings of the audio and video features, $\tau$ is a temperature parameter.  To leverage the full power of narrations on Ego4D, we initialize the video encoder weights from models pre-trained on video and language from~\cite{lin22egocentric}.

\vspace{-0.15in}
\subsection{Implementation Details}
\vspace{-0.05in}

We use Ego4D-Sounds (see \cref{sec:ego4dsounds}) to train our AV-LDM. Video is sampled at 5FPS and audio is sampled at 16kHz. Video is passed through the pre-trained video encoder to produce condition features $c_v\in \mathbb{R}^{16\times 768}$. The audio waveform is transformed into a mel-spectrogram with a hop size of 256 and 128 mel bins. The mel-spectrogram is then passed to the VAE encoder with padding in the temporal dimension to produce target $z_0\in\mathbb{R}^{4\times 16\times24}$. The audio condition is processed the same way except that we use an additional MLP to process VAE's output to produce $c_a \in \mathbb{R}^{24\times 768}$. We load the weights of VAE and LDM from the pretrained Stable Diffusion to speed up training, similar to~\cite{luo23difffoley}, and VAE is kept frozen during training. LDM is trained for $8$ epochs with batch size $720$ on Ego4D-Sounds with the AdamW optimizer with learning rate $1e-4$. During inference, we use $25$ sampling steps with classifier-free guidance scale $\omega=6.5$. For HiFi-GAN, we train it on a combination of 0.5s segments from Ego4D\cite{grauman22ego4d}, Epic-Kitchens~\cite{huh23epicsounds}, and AudioSet~\cite{gemmeke17audioset}. We use AdamW to train HiFi-GAN with a learning rate of $2e-4$ and batch size of $64$ for 120k steps. We set the number of random nearby audio samples $X=6$. See more details in Supp.
\vspace{-0.15in}
\section{The Ego4D-Sounds Dataset}
\label{sec:ego4dsounds}
\vspace{-0.05in}

\begin{figure}[t]
    \centering
    \includegraphics[width=\linewidth]{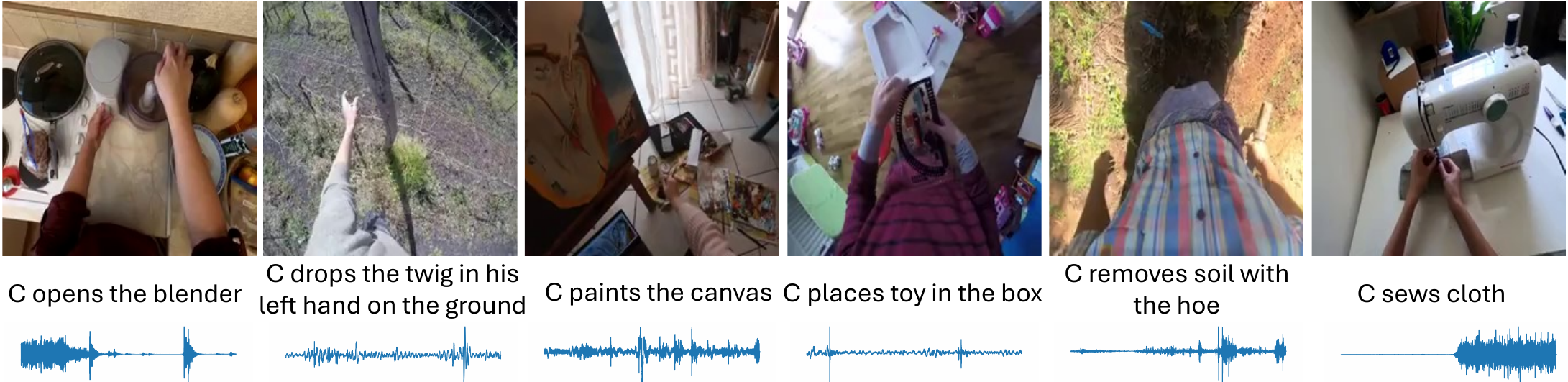}
    \vspace{-0.25in}
    \caption{Example clips in Ego4D-Sounds. We show one video frame, the action description, and the sound for each example. Note how these actions are subtle and long-tail, usually not present in typical
    video datasets.}
    \label{fig:dataset_distribution}
    \vspace{-0.1in}
\end{figure}

\begin{table}[t]
\setlength{\tabcolsep}{5pt}
    \centering
    \begin{tabular}{c|c|c|c}
    \toprule
    Datasets     & Clips & Language & Action Types \\
    \midrule
    The Greatest Hits~\cite{owens2016visually}   & 46.6K & \xmark & Hit, scratch, prod\\
    VGG-Sound~\cite{chen20vggsound}           & 200K  & Video tags & Not action-specific \\
    EPIC-SOUNDS~\cite{huh23epicsounds}         & 117.6K  & Audio labels & Kitchen actions\\
    Ego4D-Sounds                            & 1.2M  & Action narrations & In-the-wild actions\\
    \bottomrule
    \end{tabular}
    \caption{Comparison with other audio-visual action datasets. Ego4D-Sounds not only has one order of magnitude more clips, but it is also coupled with language descriptions, supporting evaluation of sound generation based on semantics.
    }
    \label{tab:dataset_comparison}
    \vspace{-0.3in}
\end{table}

Next we describe how we curate Ego4D-Sounds, an audio-video dataset for human action sound generation. 
Our goal is to curate a \CAcr{large-scale} high-quality dataset for action-audio correspondence for action-to-sound generation, addressing the issue of limited action types \CAcr{and scale} in the existing impact sound datasets~\cite{owens2016visually,clarke2023realimpact}, \CAcr{as well as more general audio-video datasets~\cite{chen20vggsound,huh23epicsounds}.} 

Ego4D~\cite{grauman22ego4d} is an existing large-scale egocentric video dataset that has more than 3,600 hours of video recordings depicting hundreds of daily activities; 2,113 of those hours have audio available. It also has time-stamped narrations that are free-form sentences describing the current activity performed by the camera-wearer. 
Utilizing the narration timestamps in Ego4D to extract clips \CAcr{directly results in a noisy dataset, since} not all clips have meaningful action sounds and there are many actions like ``talk with someone'', ``look around'', ``turn around'' that have low audio-visual correspondence. \CAcr{To ensure Ego4D-Sounds has high action-sound correspondence,} we use an automatic pipeline \CAcr{that consists of metadata-based filtering, audio tagging, and energy-based filtering} to process all extracted clips, which yields 1.2 million audio-visual action clips plus 11k clips for evaluation. See Supp. for more details on the data processing pipeline.
We show examples in \cref{fig:dataset_distribution} and comparison with other datasets in \cref{tab:dataset_comparison}. 

For all resulting clips, we extract them as 3s clips with $224\times 224$ image resolution at $30$ FPS. For audio, we extract them as a single channel with a $16000$ sample rate. 
\vspace{-0.15in}
\section{Experiments}
\vspace{-0.05in}
\label{sec:experiments}

To evaluate the performance of our model, we use the following metrics:
\begin{enumerate} 
\vspace{-0.1in}
    \item Fréchet Audio Distance (FAD)~\cite{fad}: evaluates the quality of generated audio clips against ground truth audio clips by measuring the similarity between their distributions.
    We use the public pytorch implementation.\footnote{\url{https://github.com/gudgud96/frechet-audio-distance}}
    \item Audio-visual synchronization (AV-Sync)~\cite{luo23difffoley}: a binary classification model that classifies whether the video and generated audio streams are synchronized. Following~\cite{luo23difffoley}, we create negative examples by either shifting audio temporally or sampling audio from a different video clip. Details in Supp.
    \item Contrastive language-audio contrastive (CLAP) scores~\cite{laionclap2023}: evaluates the semantic similarity between the generated audio and the action description. We finetune the CLAP model\footnote{\url{https://github.com/LAION-AI/CLAP}} on the Ego4D-Sounds data and compute scores for the generated audio and the narration at test time.
    
\vspace{-0.1in}
\end{enumerate}
These metrics measure different aspects of generation collectively, including the distribution of generated samples compared to the ground truth clips, synchronization with the video, and the semantic alignment with the action description.

We compare with the following baseline methods: 
\begin{enumerate}
\vspace{-0.1in}
    \item Retrieval: we retrieve the audio from the training set using the $\text{AV-Sim}$ model introduced in \cref{sec:pretrain}. This method represents retrieval-based generation models such as ImageBind~\cite{girdhar2023imagebind}.
    \item REGNET~\cite{chen2020regnet}: \CAcr{a video-to-audio model that uses a bottleneck design to generate visually-relevant sounds. We run their trained model on our test set.}
    \item Spec-VQGAN~\cite{iashin21taming}: a video-to-audio model that generates audio based on a codebook of spectrograms. We run their pre-trained model on our test set.
    \item Diff-Foley~\cite{luo23difffoley}: a recent LDM-based model. We follow their fine-tuning steps on egocentric videos to train on our dataset.
    \vspace{-0.1in}
\end{enumerate}


In addition, we provide ablations: ``w/o vocoder'': we replace the trained HiFi-GAN vocoder with Griffin-Lim; ``w/o cond'': we remove the audio condition at training time; ``w/o cond + denoiser'': we use an off-the-shelf model to denoise the target audio~\footnote{\url{https://github.com/timsainb/noisereduce}};  ``w/ random test cond'': we use random audio from the training set as the condition instead of retrieving audio with the highest $\text{AV-Sim}$ score.

\begin{table}[t]
\setlength{\tabcolsep}{5pt}
\centering
\begin{tabular}{ccccc}
\toprule
 & FAD $\downarrow$ & AV-Sync (\%)$\uparrow$ & CLAP$\uparrow$ \\
\midrule
Ground Truth (Upper Bound) & 0.0000 & 77.69 & 0.2698  \\

Retrieval           & 1.8353 & 11.84 & 0.0335   \\
REGNET~\cite{chen2020regnet} & 8.3800 & 3.90 & 0.9900 \\
Spec-VQGAN~\cite{iashin21taming} & 3.9017 & 7.12 & 0.0140   \\
Diff-Foley~\cite{luo23difffoley} & 3.5608 & 5.98 & 0.0346  \\
\midrule
Ours w/o vocoder    & 4.9282 & 29.60 & 0.1319 \\
Ours w/o cond +  denoiser    & 1.4676 & 1.09 & 0.0009 \\
Ours w/o cond       & 1.4681 & 39.63 & 0.1418  \\
Ours w/ random test cond & 1.0635 & 28.74 & 0.1278   \\
AV-LDM (Ours)              &\B 0.9999 &\B 45.74 &\B 0.1435 \\
 \bottomrule
\end{tabular}
\caption{Results on Ego4D-Sounds test set. 
We also report the performance of the ground truth audio, which gives the upper bound value for each metric. 
}
\label{tab:main_results}
\vspace{-0.35in}
\end{table}

\begin{figure}[t]
    \centering
    \includegraphics[width=\linewidth]{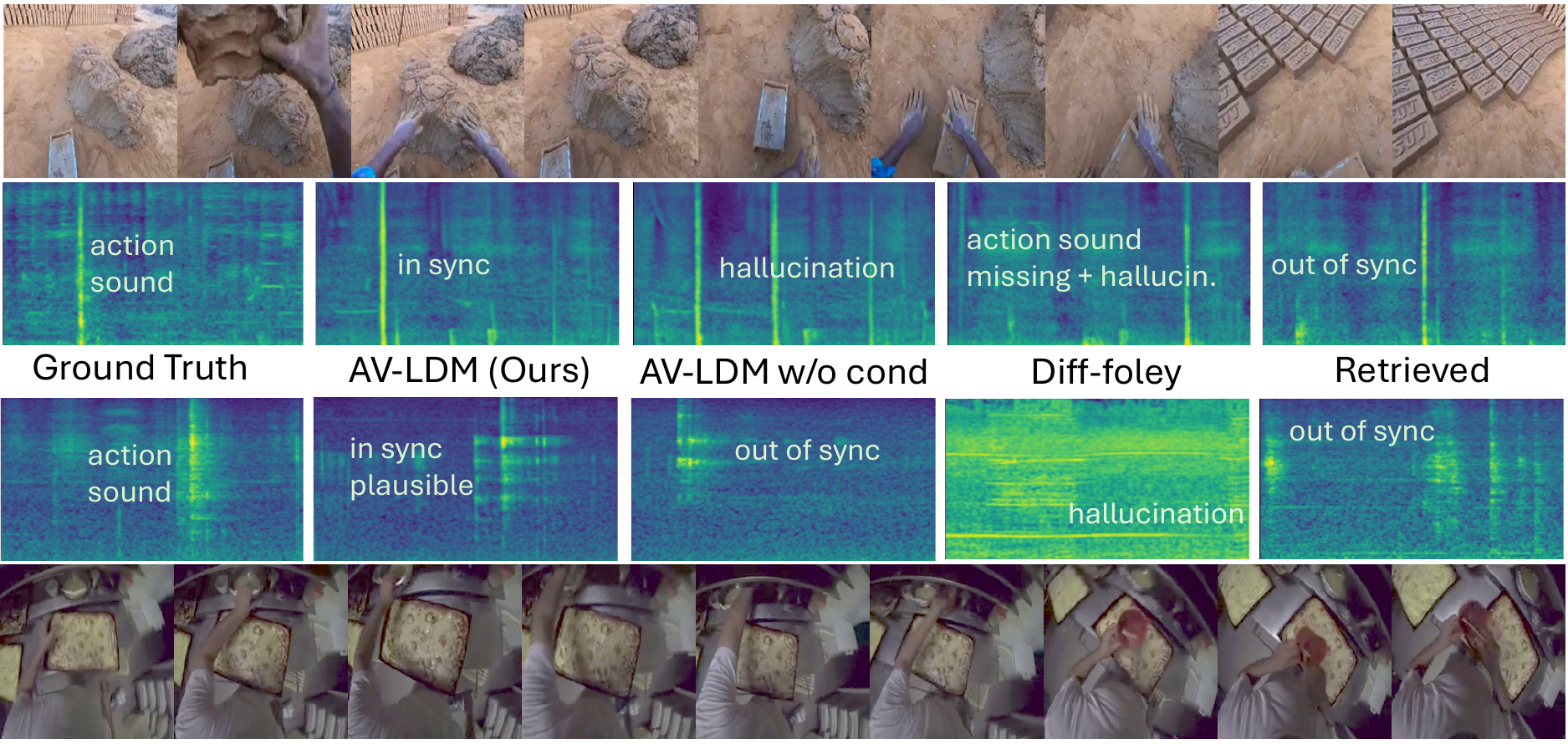}
        \vspace{-0.25in}
    \caption{Qualitative examples showing frames followed by the waveform/spectrogram of various baselines. 
    Our model generates the most synchronized sounds. 
    }
    \label{fig:main_qual}
        \vspace{-0.2in}
\end{figure}


\vspace{-0.1in}
\subsection{Results on Ego4D-Sounds}
\vspace{-0.05in}
First we evaluate the ambient-sound joint generation setting with retrieval augmented generation.
\cref{tab:main_results} shows the results.  Compared to all three baselines, we outperform them on all three metrics by a large margin. While the Retrieval baseline retrieves natural sounds from the training set and has a low FAD score compared to Spec-VQGAN and Diff-Foley, both its AV-Sync accuracy and CLAP scores are very low. Diff-Foley performs better than Spec-VQGAN since it has been trained on this task, but it still largely underperforms our model w/o cond, likely because its video features do not generalize to the egocentric setting well. 
\CAcr{REGNET performs the worst likely due to its failure to account for strongly present ambient sounds and its assumption for a fixed taxonomy of sounds.}  

For ablations, ``Ours w/o cond" has a much worse FAD score compared to the full model, showing the importance of our ambient-aware training. As expected, ``Ours w/o cond + denoiser" has very low scores on AV-Sync and CLAP since existing noise reduction algorithms are far from perfect.
We also test our model by conditioning it on a random audio segment at test time instead of the one retrieved with the highest audio-visual similarity and its performance also gets worse, verifying the effectiveness of our retrieval-based solution.

We show two qualitative examples in \cref{fig:main_qual}  
comparing our model with several baselines.  Our model synthesizes both more synchronized and more plausible sounds. 
To fully evaluate our results, it is important to view the Supp. video.

\begin{figure}[t]
  \centering
  \begin{subfigure}[b]{0.49\textwidth}
    \includegraphics[width=\textwidth]{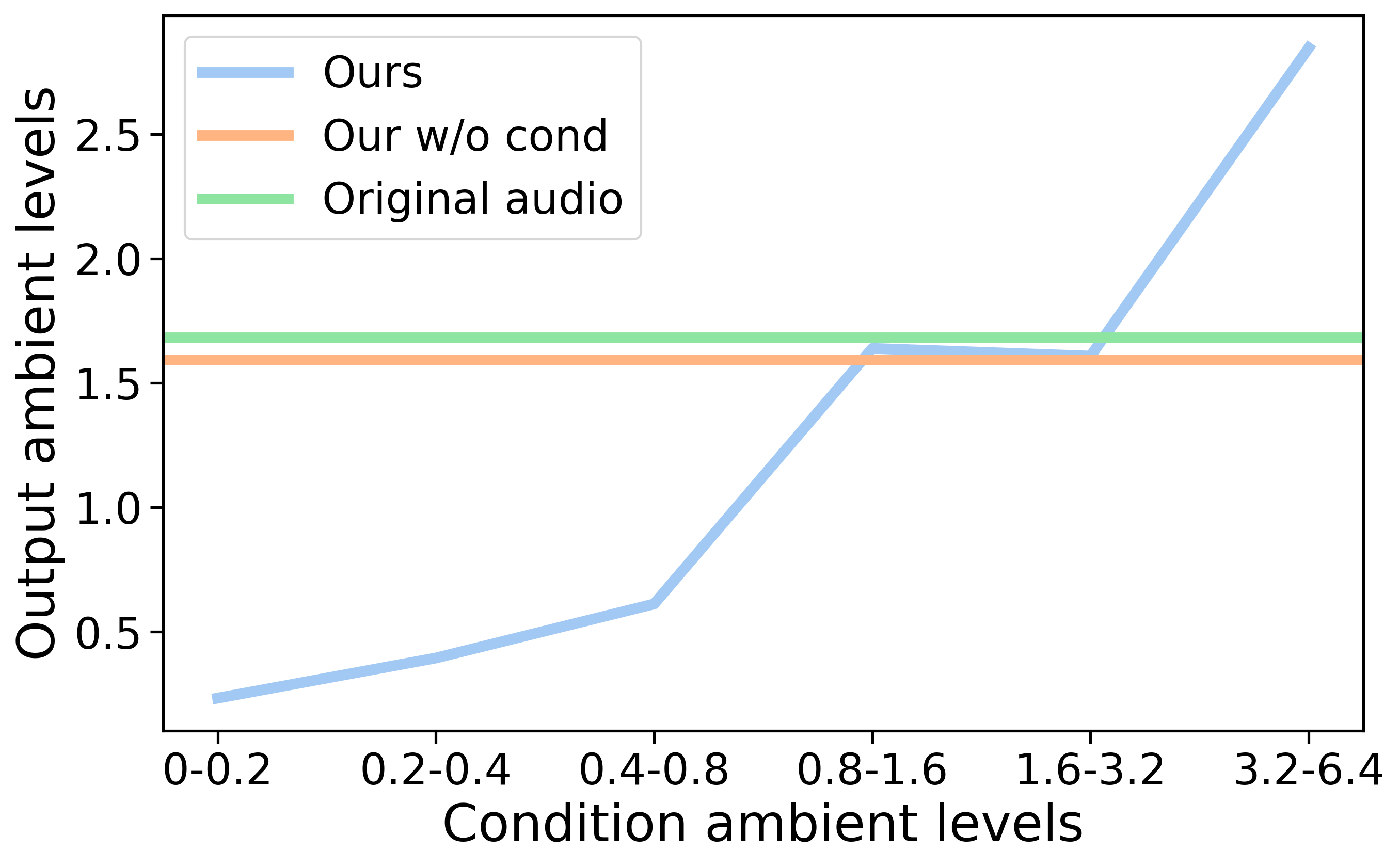}
    \caption{Varying ambient level condition}
    \label{fig:varying_ambient_ambient}
  \end{subfigure}
  \hfill 
  \begin{subfigure}[b]{0.49\textwidth}
    \includegraphics[width=\textwidth]{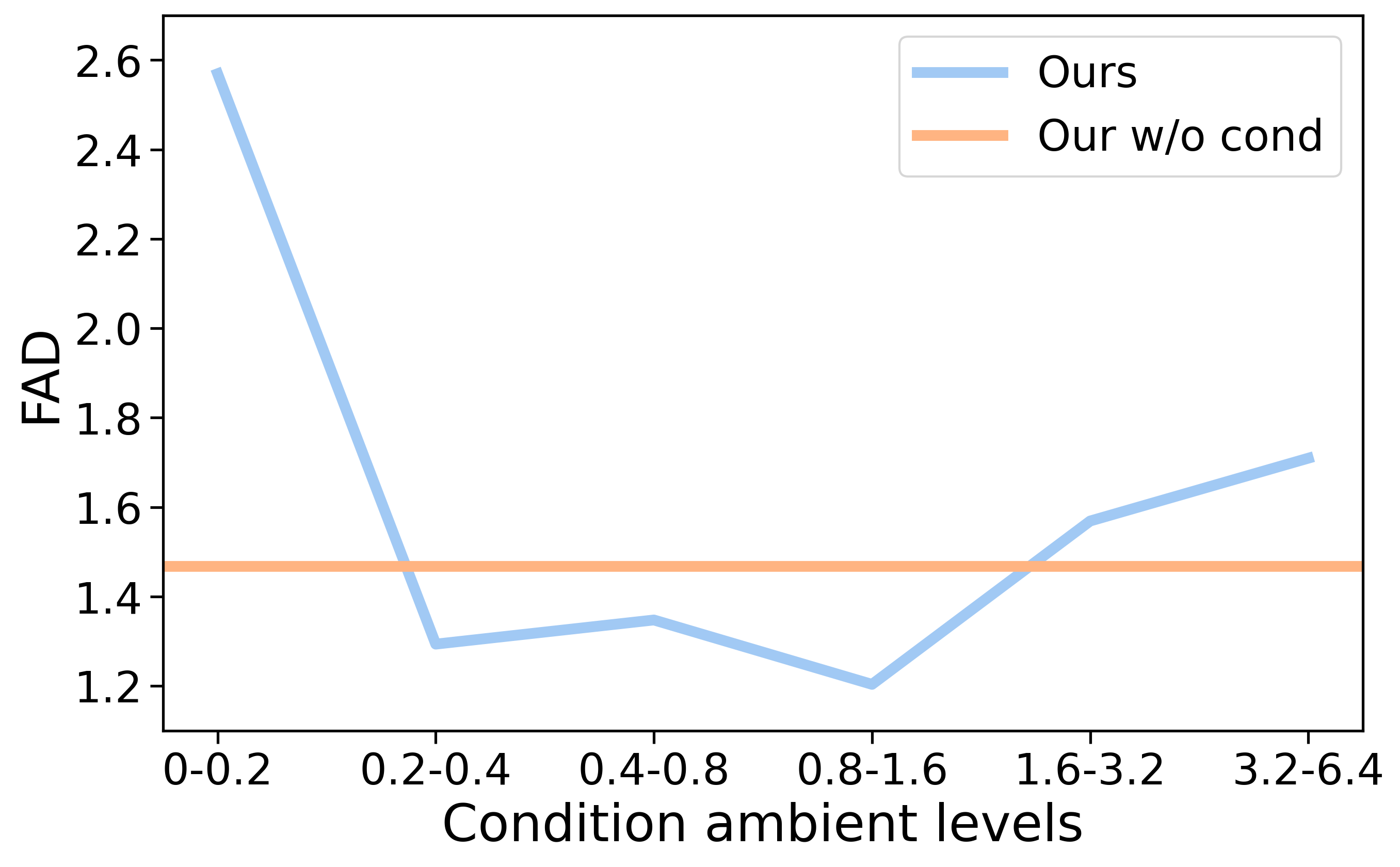}
    \caption{Audio generation accuracy (FAD)}
    \label{fig:varying_ambient_fad}
  \end{subfigure}
      \vspace{-0.1in}
  \caption{The achieved ambient level and accuracy 
   as a function of the input ambient levels. (a) We show the ambient level of our model changes according to the ambient level in the audio condition while the ambient level of ``Ours w/o cond'' and the original audio stay constant, illustrating the controllability of our model. (b) FAD is low for most input ambient levels unless it goes too extreme (too low or too high), showing our model generates high-quality action sounds even when varying output ambient levels.
  }
  \label{fig:varying_ambient}
      \vspace{-0.1in}
\end{figure}

\vspace{-0.15in}
\subsection{Ambient Sound Control}\label{sec:ambient_control}
\vspace{-0.05in}

By disentangling action sounds from ambient sounds, our model allows taking any given sound as the condition at test time.
To examine whether our model truly relies on the audio condition to learn the ambient sound information, we next test the model by providing audio conditions of various ambient levels and then calculate the ambient level in the generated audio. The ambient level is defined as the lowest energy of any 0.5s audio segment in a 3s audio.

\cref{fig:varying_ambient} shows the results, where we also plot the ambient levels of ``Ours w/o cond'' and the original audio.
Our model changes the ambient sound level according to the input ambient (shown in \cref{fig:varying_ambient_ambient}) while still synthesizing plausible action sounds (shown in \cref{fig:varying_ambient_fad}). FAD spikes when the condition ambient is too low or too high, most likely because the generated ambient sound is out of distribution since the original audio always has some ambient sounds.

\cref{fig:ambient_comparison} shows example outputs from our model and several baselines. The examples
show how our model generates plausible action sounds when conditioned on a low-ambient sound for action-focused generation. We can see that the action-focused setting generates similar action sounds as the action-ambient setting while having a minimal ambient level. While by definition we lack a good evaluation of this setting (there is no ground truth audio source separation for the data), our model shows an emerging capability of generating clean action sounds although it has never been explicitly trained to do so.

\begin{figure}[t]
    \centering
        \vspace{-0.1in}
    \includegraphics[width=\linewidth]{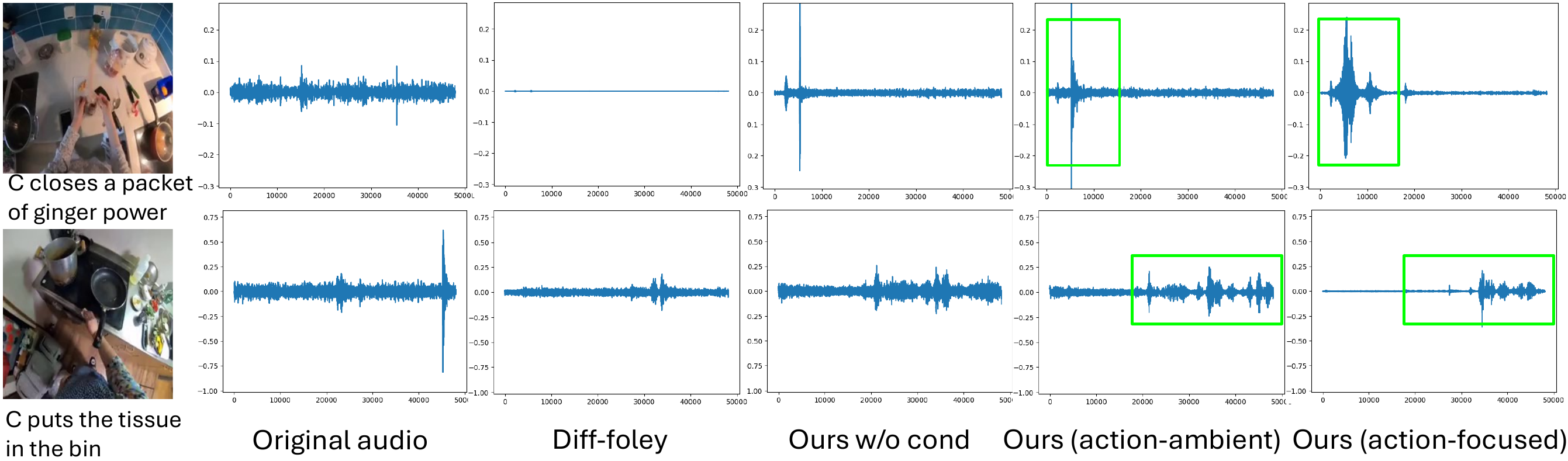}
    \vspace{-0.2in}
    \caption{Visualization of action-focused generation. For both examples, Diff-Foley~\cite{luo23difffoley}, Ours w/o cond or Ours (action-ambient generation) generate plausible action sounds along with ambient sounds. In contrast, our model conditioned on a low ambient sound generates plausible action sounds (see green boxes) with minimal ambient sound.
    } 
    \label{fig:ambient_comparison}
    \vspace{-0.1in}
\end{figure}

\vspace{-0.15in}
\subsection{Human Evaluation}
\vspace{-0.05in}
To further validate the performance of various models, we conduct a subjective human evaluation.
In each survey, we provide 30 questions, each with 5 videos with the same visuals but different audio samples. For each video, we ask the participant to select the video(s) whose audio 1) is most semantically plausible and temporally synchronized with the video and 2) has the least ambient sounds. We invite 20 participants to complete the survey and compute the average voting for all 30 examples. See the survey interface and guidelines in Supp.

\cref{tab:human_evaluation} shows the results. All learning-based methods generate reasonable action sounds, yet our model (action-ambient) has the highest score for action-sound quality compared to other methods. Although ours (action-focused) has a slightly lower action-sound score, it has significantly less ambient sound. This is likely because sometimes the low-ambient condition can lead the model to suppress some minor action sounds. 

\CAcr{Overall, our model generates both short percussive and longer harmonic action sounds while producing desired ambient sounds controlled by users. The model can fail sometimes in predicting more subtle action sounds, however.
See Supp. video for both success and failure examples.}


\begin{table}[t]
    \centering
    \begin{tabular}{c|c|c}
    \toprule
                & Action sound quality &  Least ambient sound\\
    \midrule
    Retrieval   &  12.5\% & 12.5\%  \\
    Diff-Foley \cite{luo23difffoley}  &  47.5\%  & 12.5\% \\
    AV-LDM w/o cond & 55.0\%  & 17.5\%  \\
    AV-LDM (action-focused) & 60.0\%  &\B  97.5\% \\
    AV-LDM (action-ambient) &\B 72.5\%  &  22.5\% \\
    \bottomrule
    \end{tabular}
    \caption{Survey results showing user preferences.  Higher is better. 
    Our model in the action-ambient joint generation setting scores highest for action sound quality, showing its ability to produce action-relevant sounds despite training with in-the-wild data. Ours in the action-focused generation setting scores highest for the least ambient sound, at a slight drop in action sound quality score, showing the ability to eliminate background sounds when requested by the user.
    }
    \label{tab:human_evaluation}
        \vspace{-0.3in}
\end{table}

\vspace{-0.15in}
\subsection{Results on EPIC-KITCHENS}
\vspace{-0.05in}

To evaluate whether our model generalizes to other datasets, we also test our model on the EPIC-KITCHENS dataset. 
We first sample 1000 3s clips from EPIC and then 
evaluate the retrieval baseline, Diff-Foley, Ours w/o cond, and our full model on these data and then compute their FAD and AV-Sync scores. 

\cref{tab:epic} shows the results. Similar to what we observe on Ego4D-Sounds, our model outperforms other models by a large margin, showing it better learns to generate action sounds from visuals, even when transferring to another dataset.

\begin{table}[t]
\setlength{\tabcolsep}{5pt}
\centering
\begin{tabular}{cccccc}
\toprule
 &GT&Retrieval&Diff-Foley  & Ours w/o cond & AV-LDM (Ours)  \\
\midrule
FAD $\downarrow$ &0.0000&1.9618&3.4649&1.4731& \B1.3200\\
AV-Sync (\%) $\uparrow$ &73.94&13.84&14.19&50.42& \B59.26\\
 \bottomrule
\end{tabular}
\caption{Results on EPIC-KITCHENS. 
GT stands for Ground Truth.}
\label{tab:epic}
\vspace{-0.3in}
\end{table}

\vspace{-0.15in}
\subsection{Demo on VR Cooking Game}
\vspace{-0.05in}

One compelling application of action-to-sound generation is to generate sound effects for games in virtual reality, where simulating complex hand-object interactions is non-trivial. To examine whether our learned model generalizes to VR games, we collect game videos of a cooking VR game ``Clash Of Chefs'' from YouTube and test our model without fine-tuning.  Preliminary results suggest our model can generate synced action sounds 
(see \cref{fig:vr_game} and Supp). 
Though there remains much work to do, this suggests a promising future in learning action-to-sound models from real-world egocentric videos and applying them to VR games to give a game user an immersive audio-visual experience that dynamically adjusts to their own actions.

\begin{figure}[t]
    \centering
    \includegraphics[width=0.8\linewidth]{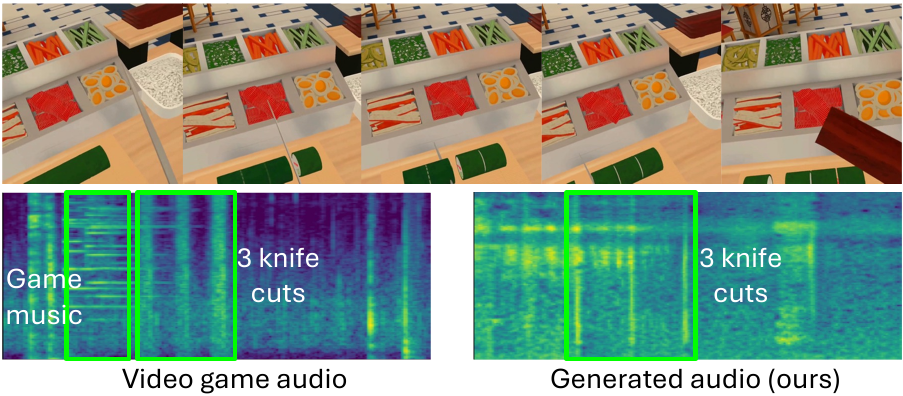}
        \vspace{-0.1in}
    \caption{We apply our model on a VR cooking game clip where the person cuts a sushi roll three times. Our model successfully predicts the 3 cutting sounds.}
    \label{fig:vr_game}
        \vspace{-0.2in}
\end{figure}
\vspace{-0.15in}
\section{Conclusion}
\label{sec:conclusion}
\vspace{-0.1in}

We investigate the problem of generating sounds for human actions in egocentric videos. 
We propose an ambient-aware approach that disentangles the action sound from the ambient sound, allowing successful generation after training with diverse in-the-wild data, as well as controllable conditioning on ambient sound levels.
We show that our model outperforms 
existing methods and baselines---both quantitatively and through human subject studies.  Overall, it significantly broadens the scope of relevant training sources for achieving action-precise sound generation.
In future work we aim to explore the possibilities for sim2real translation of our learned audio generation models to synthetic imagery inputs, e.g., for VR game applications.


\CAnote{I didn't find a proper place to insert discussion on this}

\CAnote{we did experiments on this but we did not save the exact numbers in the past so I did not include it.}
\CAnote{added in supp, last section}
\CAnote{added in sec 3.3, which I think explains Q2 as well, that is this applies to both training and evaluation}

\paragraph{Acknowledgments:} UT Austin is supported in part by the IFML NSF AI Institute. Wei-Ning Hsu helped advise the project only, and all the work and data processing were done outside of Meta.

\bibliographystyle{splncs04}
\bibliography{changan_general,pyp,changan_specific}

\clearpage
\setcounter{section}{6}
\section{Supplementary}

In this supplementary material, we provide additional details about:
\begin{enumerate}
    \item Supplementary video for qualitative examples (referenced in Sec.~1).
    \item Additional implementation details (referenced in Sec.~3).
    \item Dataset details (referenced in Sec.~4).
    \item Evaluation metric details (referenced in Sec.~5).
    \item Human evaluation details  (referenced in Sec.~5).
    \item Effect of the size of the retrieval pool.
    \item Ablation study on video frame rates.
    
\end{enumerate}

\subsection{Supplementary Video}
In this video, we include examples of Ego4D-Sounds clips, qualitative examples on unseen Ego4D clips, and qualitative examples on VR games. Wear headphones to hear the sound.

\subsection{Additional Implementation Details}
\noindent\textbf{Audio-Visual LDM.} Our Ego4D-Sounds clips are 3 seconds long. For model training and inference, we sample audio waveform at 16000Hz. We use FFT size 1024, mel bins 128, hop size 256 to transform the 3-second audio waveform into a mel-spectrogram of length 188, which we then pad in the temporal dimension to $192$.
To speed up training, similar to~\cite{luo23difffoley}, we load VAE and diffusion model weights from the pre-trained Stable Diffusion model. Note that Stable Diffusion expect image as the input/target, and therefore we duplicate the mel-spectrogram in the channel dimension and to achieve size $x_0 \in \mathbb{R}^{3\times 128 \times 192}$, passing $x_0$ to the VAE encoder, we get compressed latent representation $z_0 \in \mathbb{R}^{4\times 16\times 24}$. For conditioning, videos are sampled at 5 FPS, passed through the video encoder and a linear projection layer that produces features of size $c_v \in \mathbb{R}^{16\times 768}$.
Audio condition is also a 3-second clip and is processed the same way as the target audio to get $\Tilde{c}_a \in \mathbb{R}^{4\times 16\times 24}$, it is then projected to a $2$ dimensional tensor of shape $c_a \in \mathbb{R}^{24\times 768}$. For classifier-free guidance, we set the scale $\omega=6.5$, and use DPM-Solver~\cite{lu2022dpm} for accelerated inference using only $25$ sampling steps.  
For the mel-spectrogram to waveform vocoder HiFi-GAN~\cite{kong2020hifi}, we train the model from scratch with the mel-spectrogram processing hyperparameters aligned with that of our AV-LDM. During training, we freeze the pre-trained VAE, and train the LDM model on Ego4D-Sounds for 8 epochs with batch size 720. We use the AdamW optimizer with a learning rate of $1e-4$. HiFi-GAN is trained on a combination of 0.5s clips from Ego4D\cite{grauman22ego4d}, Epic-Kitchens~\cite{huh23epicsounds}, and AudioSet~\cite{gemmeke17audioset}. We use AdamW to train HiFi-GAN with a learning rate of $2e-4$ and a batch size of 64 for 120k steps.

\textbf{Audio-visual representation learning.} We use Timesformer~\cite{bertasius21spacetime} as the video encoder, and AST~\cite{gong21ast} as the audio encoder. We train video and audio encoders for 5 epochs with batch size $256$. We use the InfoNCE~\cite{infonce} loss and Adam optimizer~\cite{kingma15adam} with a learning rate $1e-4$. 

\subsection{Dataset Details}

To evaluate the effectiveness of our proposed ambient-aware action sound generation model, we leverage Ego4D~\cite{grauman22ego4d}, a large-scale egocentric video dataset for daily human activities. While our model is capable of disentangling action sound from ambient sound, there is little value in learning on data that only contain ambient sounds or background speech. Our goal is to curate an in-the-wild action dataset that has meaningful action sounds.  We design a four-stage pipeline consisting of both learning-based tagging tools and rule-based filters to curate the Ego4D-Sounds dataset. To be consistent with the public splits of Ego4D benchmarks, we use all 7.5K videos in the training set, where we extract 3.8M clips centered at the narrations' timestamps with the left and right margins being 1.5s, i.e. each clip is 3s long, \CA{which we find to be sufficiently long enough to capture the narrated action.}

We first remove all clips without sounds, resulting in 2.5M clips. We then filter the above clips based on the scenarios. Each Ego4D video has a scenario label, categorizing the activity depicted in the video. We go through all scenario labels and pick 28 scenarios that are mainly social scenarios, e.g., "playing board games", "attending a party", "talking with friends", where majority of the sounds are speech or off-screen sounds with no on-screen actions. We remove videos with these tags, resulting in 3.1K videos and 1.7M clips. 

While the previous stage removed videos for social scenarios as a whole, there are still many clips that have only speech or background music. To detect these clips, we use an off-the-shelf audio tagging tool to tag the remaining clips. The goal is to remove clips that have solely off-screen sounds, i.e., speech and music. So we threshold the tagged probability at $0.5$, i.e., removing clips that most likely only contain off-screen sounds and not action sounds.
This filtering process further removed 0.5M clips, with 1.23M clips remaining.

Lastly, we also observe that in a long video clip, there are silent periods when no sounding action occurs. Based on this observation, we devise an energy-based filtering process, i.e. we normalize the amplitude of each clip with respect to the maximum amplitude of audio in the video. We then convert the amplitude to dB and remove clips with energy below -60 dB. This results in 1.18M clips.




\subsection{Evaluation Metric Details}
For the audio-visual synchronization (AV-Sync)~\cite{luo23difffoley} metric, we train a synchronization binary classification model on Ego4D-Sounds, and use it to judge whether the generated audio is synchronized with the video. Following~\cite{luo23difffoley}, to construct the input to the classification model, we input paired and synced video and audio, unpaired video and audio, and paired but unsynced video and audio 50\%, 25\%, and 25\% of the time respectively. The model uses Timesformer~\cite{bertasius21spacetime} as the video encoder, AST~\cite{gong21ast} as the audio encoder, and a 3 layer MLP as the classification head which takes the CLS tokens from the two encoders, concatenates them in the feature dimension, and produces a probability indicating the synchronization. We train this model for 30k steps with AdamW optimizer, which achieves a classification accuracy of 70.6\% on the validation set.
    
\subsection{Human Evaluation Details}
For the human evaluation, we first compile a guideline document, clarifying and defining the objectives of the survey and what the participant should be looking for. There are two main objectives: 1) select video(s) with the most plausible action sounds (e.g., object collisions, water running) that are semantically and temporally matching with the visual frames, and 2) select video(s) with the least ambient noise. We also provide multiple positive and negative examples for each criterion in the guideline document. We ask participants to read the guidelines before doing the survey. We show one example of the survey interface in \cref{fig:survey}.

\begin{figure}[t]
    \centering
    \includegraphics[width=\linewidth]{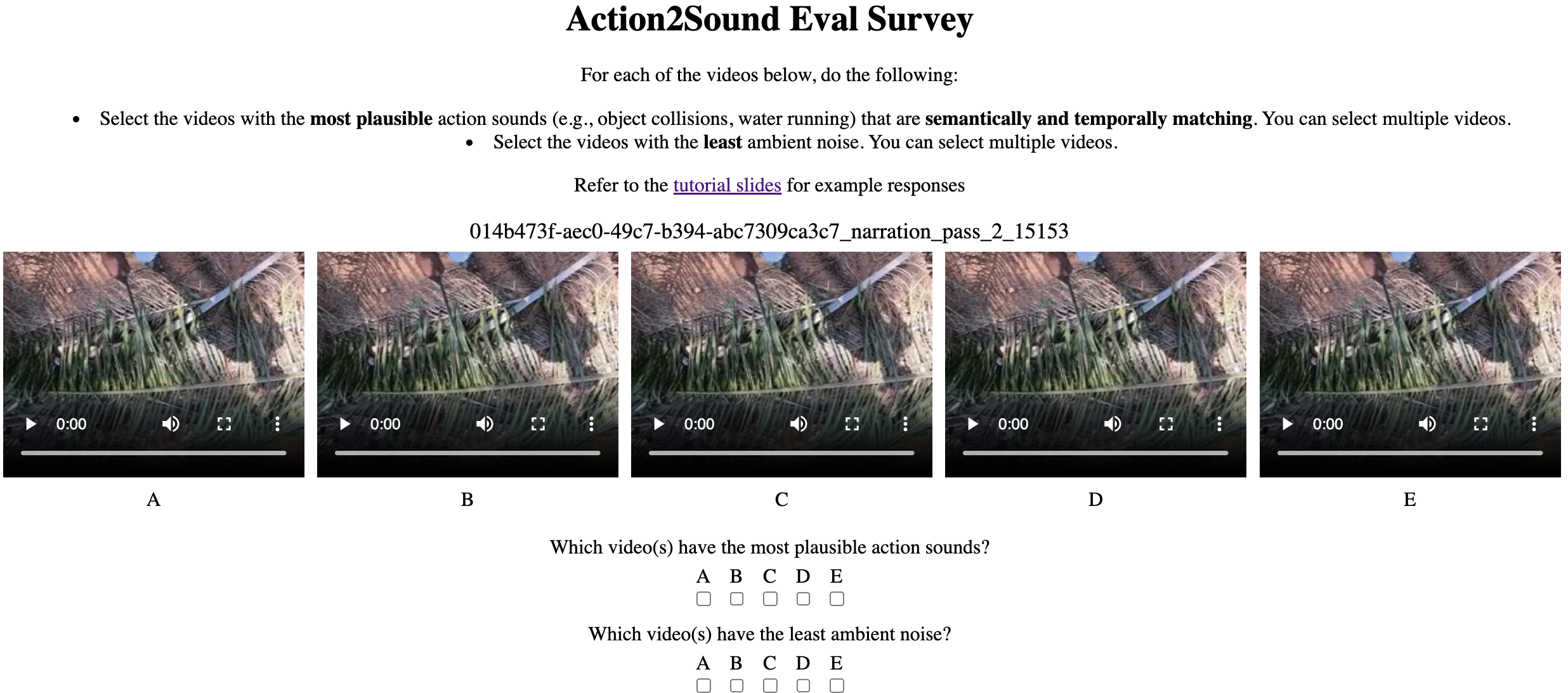}
    \caption{Survey interface. For each example, we ask participants to select video(s) that are most plausible with the video and the video(s) that have the least ambient sound.}
    \label{fig:survey}
\end{figure}

\subsection{Effect of the Size of the Retrieval Pool}
To understand how the size of the retrieval pool affects the model performance, we evaluate our model with varying pool sizes (10k/1k/100 randomly sampled from the full retrieval pool, and the FAD scores are 1.01/1.02/1.12 (lower is better). We see that the model is fairly robust to the retrieval pool size.

\subsection{Ablation Study on Video Frame Rates.}
In the paper, we use 16 frames for 3s video (around 5 FPS). We experimented with higher FPS (8 and 10) previously and did not observe significant improvements, likely because most human actions do not have higher frequencies and can be captured with 5 FPS. 
\end{document}